\documentclass[runningheads]{llncs} 
\usepackage[utf8]{inputenc}
\RequirePackage{amsmath}
\usepackage{subcaption}
\RequirePackage{graphicx}
\RequirePackage{booktabs}
\usepackage{tabularx}
\usepackage{textcomp}

\begin{document}
\title{Progressive Learning with Anatomical Priors for Reliable Left Atrial Scar Segmentation from Late Gadolinium Enhancement MRI}
\titlerunning{Atrial Scar Segmentation}
\authorrunning{J. Zhang et al.}

\author{Jing Zhang\inst{1}
\and
Bastien Bergere\inst{1}
\and 
Emilie Bollache\inst{1}
Jonas Leite\inst{1}
\and
Mikaël Laredo\inst{1,2}
\and
Alban Redheuil\inst{1,3}
\and 
Nadjia Kachenoura\inst{1}\\
}

\institute{1. Laboratoire d’Imagerie Biomédicale (LIB), Sorbonne Université, INSERM, CNRS, Paris, France
\\
2. APHP, Institut de Cardiologie, Hôpital Pitié-Salpêtrière, Paris, France\\
3. APHP, Imagerie Cardiovasculaire et thoracique, Institut de Cardiologie, Hôpital Pitié Salpêtrière, Paris, France
}

\maketitle       

\begin{abstract}
\textbf{Background.} 
    Cardiac MRI late gadolinium enhancement (LGE) enables non-invasive identification of left atrial (LA) scar, whose spatial distribution is strongly associated with atrial fibrillation (AF) severity and recurrence. However, automatic LA scar segmentation remains challenging due to low contrast, annotation variability, and the lack of anatomical constraints, often leading to non-reliable predictions. Accordingly, our aim was to propose a progressive learning strategy to segment LA scar from LGE images inspired from a clinical workflow.
\textbf{Methods.}
    A 3-stage framework based on SwinUNETR was implemented, comprising: 1) a first LA cavity pre-learning model, 2) dual-task model which further learns spatial relationship between LA geometry and scar patterns, and 3) fine-tuning on precise segmentation of the scar. Furthermore, we introduced an anatomy-aware spatially weighted loss that incorporates prior clinical knowledge by constraining scar predictions to anatomically plausible LA wall regions while mitigating annotation bias. 
\textbf{Results.}
    Our preliminary results obtained on validation LGE volumes from LASCARQS public dataset after 5-fold cross validation, LA segmentation had Dice score of 0.94, LA scar segmentation achieved Dice score of 0.50, Hausdorff Distance of 11.84 mm, Average Surface Distance of 1.80 mm, outperforming only a one-stage scar segmentation with 0.49, 13.02 mm, 1.96 mm, repectively.
\textbf{Conclusion.}
    By explicitly embedding clinical anatomical priors and diagnostic reasoning into deep learning, the proposed approach improved the accuracy and reliability of LA scar segmentation from LGE, revealing the importance of clinically informed model design.

\end{abstract}

\section{Introduction}
\label{sec:intro}

Atrial fibrillation (AF) is the most prevalent sustained cardiac arrhythmia and affects individuals across a wide age spectrum, representing a growing global health burden \cite{chugh2014worldwide,kornej2020epidemiology}. Crucially, AF is associated with a significantly increased risk of blood clots forming in the left atrium (LA), which can lead to systemic thromboembolism and ischemic stroke \cite{wolf1991atrial}. Catheter ablation through pulmonary vein isolation (PVI) has become an established therapeutic strategy for rhythm control in symptomatic AF \cite{hindricks20212020}; yet, long-term outcome in AF remains suboptimal, and a substantial proportion of patients experience AF recurrence \cite{freedman2021world}.

Increasing evidence suggests that post-ablation structural remodeling, particularly formation and distribution of LA scar tissue, plays a critical role in predicting ablation success and AF recurrence \cite{verma2015approaches}. Cardiac magnetic resonance imaging (MRI) late gadolinium enhancement (LGE) has been recognized as a non-invasive technique to visualize LA fibrosis and ablation-induced scarring \textit{in vivo} \cite{marrouche2014association}. Accurate delineation of LA scar regions from LGE images is therefore of significant clinical interest, as it enables quantitative assessment of atrial remodeling and holds promise for improving post-ablation monitoring and individualized management of AF patients \cite{siebermair2017assessment}.  However, LA wall is thin relative to LGE spatial resolution, scar regions are even smaller and may exhibit low contrast, making both manual delineation or automated segmentation challenging.

Recently, the field of LA scar delineation from LGE images has shifted from manual to efficient semi-/automatic segmentation thanks to the development of artificial intelligence (AI) techniques. For instance, Zhuang and colleagues have pioneered multimodality learning and shape modeling to address challenges of low contrast and variable scar morphology in LGE-MRI \cite{zhuang2018multivariate}. Building on this, Li et al. introduced sophisticated deep neural networks, including the multiscale sequential network and attention-guided mechanisms, which significantly improved the LA wall and scar segmentation \cite{li2022atrialjsqnet,li2022medical}. Foundation models have also been used in this field based on sufficient amount of training data \cite{tavakoli2025scarnet}. Rather than relying solely on increasingly complex deep learning (DL) architectures, we seek to bridge clinical practice and AI modeling from a first principles learning perspective. To this end, we propose a progressive learning strategy, inspired by clinical (e.g. from a resident physician) cognitive process, in which a model first learns the ``relatively easy'' LA anatomy, then the model with learned LA features captures the spatial relationship between LA and scar patterns, and, finally, the model transitions and focuses on scar areas. Our main specific contributions are summarized as follows:

\begin{itemize}

\item We proposed a clinically inspired progressive learning framework for LA scar segmentation from LGE MRI images, which mimics the diagnostic workflow of clinicians by sequentially learning LA anatomy, spatial context, and fine-grained scar delineation. This design enables stable and structured feature learning for low contrast and highly imbalanced scar regions.

\item We introduced an anatomy-aware spatially weighted loss that embeds clinical prior knowledge by softly constraining scar predictions within anatomically plausible atrial wall regions. This formulation improves robustness to annotation noise and prevents anatomically implausible outlier predictions without altering the primary segmentation objective.

\item We developed a physics-inspired, on-the-fly 3D data augmentation strategy tailored for LGE MRI, incorporating realistic intensity perturbations, spatial deformations, and label-guided sampling to enhance generalization under limited training data.

\end{itemize}

\section{Related Work}
\label{sec:related_works}

\subsection{LA scar segmentation}

Direct LA scar segmentation remains challenging due to its narrow and irregular distribution, and sensitivity to surrounding enhanced tissues. Therefore, scar segmentation is typically achieved with the assistance of LA cavity and/or LA wall segmentation either manually or using conventional or DL methods \cite{li2022medical}. 

\textbf{Conventional methods} primarily rely on intensity characteristics and anatomical constraints. Thresholding-based methods \cite{badger2010,karim2013}, including $n$ standard deviations and full width at half maximum, assume significantly higher pixel intensity in scar than in healthy myocardium, but suffer from subjective, operator-dependent threshold selection and sensitivity to image quality as well as to contrasts induced by timing after injection and inversion time optmization. Maximum intensity projection method mitigates dependency on accurate LA cavity segmentation by projecting intensities onto the LA surface \cite{knowles2010,razeghi2020}. Graph-based methods (e.g., Markov Random Field with graph-cuts) incorporate spatial smoothness priors to handle diffuse scar patterns \cite{karim2011,karim2014}. However, these methods generally require semi-automatic initialization or manual LA wall segmentation for reliable performance \cite{karim2013}.

\textbf{DL methods} have shown promising results in cardiology including in LA scar segmentation in the setting of AF \cite{gunawardhana2025integrating}. 
Early attempts combined convolutional neural networks (CNN) with conventional graphical models, such as CNN-learned features for graph-cut optimization \cite{li2018b,li2020b}. 
Multitask learning frameworks simultaneously segment LA cavity and scar, while exploiting their intrinsic spatial relationships through shared feature representations or attention mechanisms \cite{chen2018b,yang2020,li2020a,xing2023}. 
Notably, surface projection strategies, i.e., projecting 3D scar onto 2D LA meshes, have been widely adopted to alleviate class imbalance and reduce computational complexity \cite{li2020b,li2020a}. 
Fine-tuning state-of-the-art SAM and YOLO models was applied to deal with scar label noise problem \cite{moafi2025robust}.
Recent studies have explored clinically and physiologically informed multimodal approaches for myocardial scar segmentation beyond purely image-based models. As such, Ramzan et al. \cite{ramzan2025seeing} introduced clinically guided LGE-based scar segmentation and later extended this framework by incorporating electrocardiogram (ECG) signals and anatomical priors for multimodal scar segmentation. 
Besides, the 2022 Left Atrial and Scar Quantification \& Segmentation Challenge (LASCARQS) \cite{zhuang2023left} was held to encourage participative efforts to scar segmentation. The most commonly used network architectures in this challenge were 3D UNET \cite{cciccek20163d} and nnUNET \cite{isensee2021nnu}. 
However, despite substantial advances brought by DL, current segmentation strategies still struggle with inherent complexity of scar morphology and signal heterogeneity \cite{gunawardhana2025integrating}. Instead of pursuing incremental improvements within the same aforementioned methodological advances, our work adopts a distinct perspective, aiming to narrow the performance gap.

\subsection{Progressive learning}

Progressive learning refers to a class of training strategies in which models are trained in a stage-wise manner, gradually increasing learning complexity. One form of progressive learning is curriculum learning \cite{bengio2009curriculum}, where training samples are organized from easy to hard to facilitate optimization and improve generalization. More broadly, progression can be defined over data difficulty, spatial context, model capacity, or task complexity. The central principle is a controlled transition from simple to complex learning scenarios \cite{fayek2020progressive}.

Existing studies in medical image segmentation have instantiated progressive learning along different dimensions of complexity.
At the data-difficulty level, Cui et al. \cite{cui2022deep} organized myocardial pathology segmentation training from relatively simple regions to more complex pathological areas in multi-sequence cardiac MRI, such that the model first captures stable structures before focusing on heterogeneous lesions.
Similarly, Jiang et al. \cite{jiang2022deep} incorporated curriculum learning into LGE MRI LA segmentation by introducing training samples according to segmentation difficulty, enabling a gradual transition from easier anatomical cases to more challenging atrial structures.
Progression can also be defined spatially. Fischer et al.~\cite{fischer2024progressive} proposed progressively increasing patch size during 3D segmentation training, where the model first learns from small local patches and subsequently integrates larger contextual regions, shifting from local pattern recognition to global structural understanding.
From a confidence perspective, Islam et al. \cite{islam2023paced} designed a paced curriculum distillation framework that starts with low-uncertainty samples and gradually incorporates high-uncertainty regions, guiding the model from confident predictions toward ambiguous boundaries.
To address domain variability, Liu et al. \cite{liu2021style} introduced style curriculum learning, in which training progresses from images with minor to stronger appearance shifts, such as intensity distribution, texture, and contrast.

Overall, progressive learning has been mainly explored in medical image segmentation tasks such as lesion detection, imbalanced learning, uncertainty-aware modeling, and domain adaptation. These studies indicate that structured increases in learning difficulty can improve training stability in challenging settings.
LA scar segmentation from LGE MRI is characterized by extreme class imbalance, heterogeneous enhancement, and ambiguous boundaries. We hypothesize that guiding the model from stable anatomical structures to sparse and uncertain scar regions can facilitate more effective feature learning and improve segmentation performance by enforcing a controlled transition from simple to complex patterns.

\section{Methodology}
\label{sec:methodology}

\subsection{Loss function and data augmentation}


LGE MRI faces two challenges: severe class imbalance between scar or LA wall and surrounding structures, and limited availability of annotated data due to protect regulations as well as time- and labor- intensive manual labeling. For instance, the average scar fraction is only 0.69\%, about 108,358 scar voxels out of ($576\times576\times44$) over 60 LGE volumes in the LASCARQS public dataset. To mitigate these limitations, numerous strategies have been proposed \cite{terven2025comprehensive}. In this work, we adopt compound loss functions to address class imbalance, specifically a combination of Dice loss and cross-entropy (DiceCE) loss:

\begin{equation}
\begin{aligned}
\mathcal{L}_{\text{Dice}} &= 1 - \frac{2 \sum_{i} \hat{y}_i y_i + \epsilon}{\sum_{i} \hat{y}_i + \sum_{i} y_i + \epsilon}, \\
\mathcal{L}_{\text{CE}} &= -\frac{1}{N} \sum_{i=1}^{N} \Bigl[ y_i \log \hat{y}_i + (1 - y_i) \log (1 - \hat{y}_i) \Bigr], \\
\mathcal{L}_{\text{DiceCE}} &= \lambda \cdot \mathcal{L}_{\text{Dice}} + (1-\lambda) \cdot \mathcal{L}_{\text{CE}}.
\end{aligned}
\end{equation}

where $y_i \in \{0,1\}$ is ground-truth label at voxel $i$; $\hat{y}_i \in [0,1]$ represents predicted probability; $N$ denotes total number of voxels; $\epsilon$ is smoothing constant for numerical stability; $\lambda \in [0,1]$ is the trade-off parameter between Dice loss and cross-entropy loss.

In addition, to improve model generalization and robustness to variations commonly observed in LGE imaging, we apply a comprehensive random on-the-fly data augmentation pipeline during training based on realistic intensity inhomogeneity, geometric distortions, scanner-related variations, and acquisition artifacts. Augmented image ($\mathcal{X'}$) after transforming the input image ($\mathcal{X}$) follows:

\begin{equation}
\mathcal{X'} = T_k \circ T_{k-1} \circ \cdots \circ T_1 (\mathcal{X})
\end{equation}

where each $T_k$ is an independently and randomly sampled transformation drawn from set of $k=8$ augmentations, described in Table \ref{tab:da}. Each transformation is applied with a certain probability or skipped, and transformation parameters are randomly sampled from predefined ranges whenever the operation is applied.

\begin{table}[ht]
\centering
\caption{Data augmentation operations along with rationale. Each operation has a probability (P) and range of variation.}
\label{tab:da}
\resizebox{0.99\textwidth}{!}{  
\begin{tabular}{p{1.2cm}p{2.80cm}p{2.2cm}p{0.9cm}p{4.2cm}}
\toprule
\multicolumn{2}{l}{\textbf{Operation}}& \textbf{Degree} & \textbf{P} & \textbf{Rationale} \\ 
\midrule

$T_{\text{bias}}$ &Bias Field & [0.02, 0.06] & 0.15 & Magnetic field inhomogeneity \\ 
$T_{\text{elastic}}$ &3D Elastic & $\sigma$: [3, 5]; $\Delta^*$ & 0.10 & Anatomical deformation \\ 
$T_{\text{rot}}$ &Rotation  & [-15$^\circ$, 15$^\circ$] & 0.50 & Rigid transformation \\ 
$T_{\text{zoom}}$ &Zoom & [0.9, 1.1] & 0.30 & Rigid transformation \\ 
$T_{\text{blur}}$ &{Gaussian Blurring} & [0.5, 1.5] & 0.20 & Reconstruction effect \\ 
$T_{\text{contrast}}$& Contrast  & [0.8, 1.3] & 0.30 & Reconstruction effect \\ 
$T_{\text{shift}}$ &{Intensity Shift} & 0.1 & 0.50 & Reconstruction effect \\ 
$T_{\text{noise}}$ &{Gaussian Noise} & $\mu$: 0, $\delta$: 0.02 & 0.20 & Acquisition effect \\ 
\bottomrule
\end{tabular}
}
\par\smallskip
\raggedright\noindent{\footnotesize $^*$ $\Delta_x,\Delta_y \in [3,6]$, $\Delta_z \in [2,5]$ voxels.}
\end{table}

\begin{figure}
    \centering
    \includegraphics[width=0.95\linewidth]{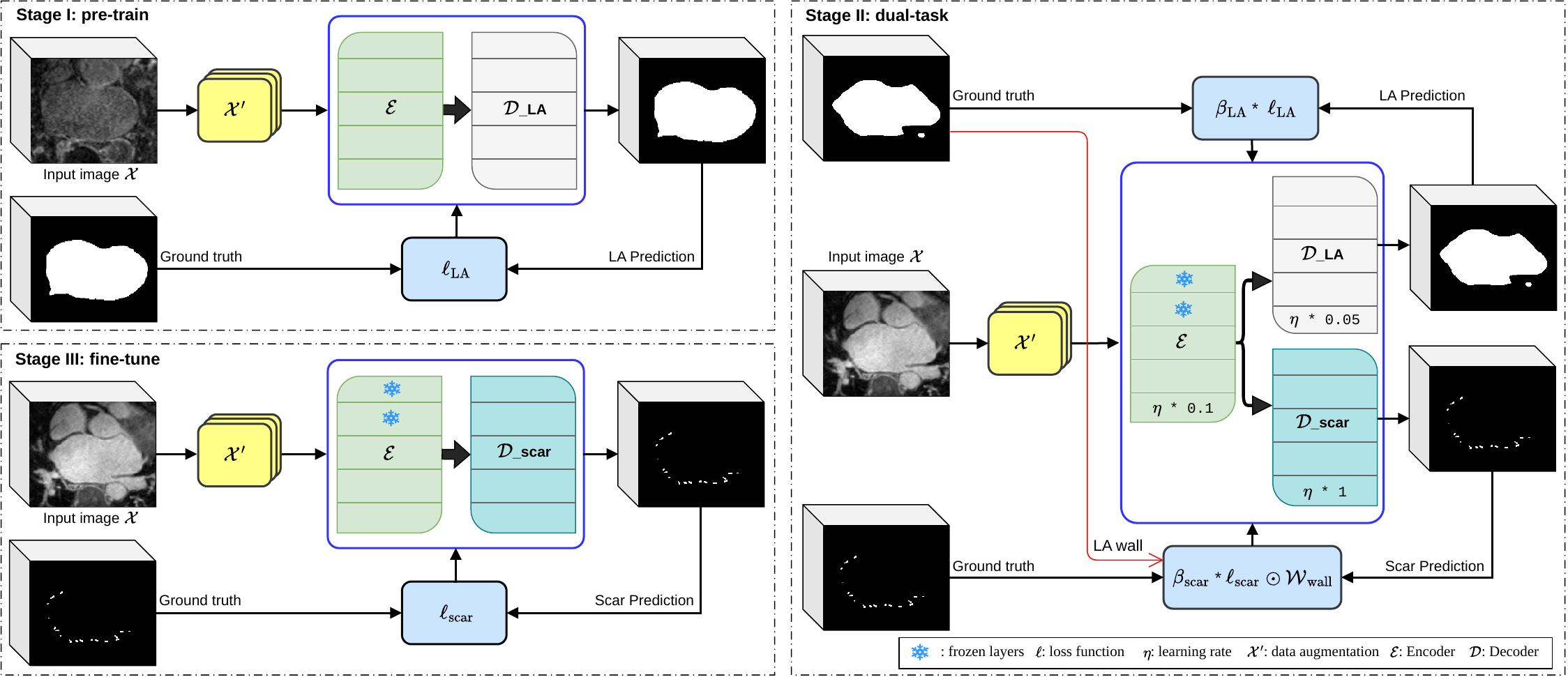}
    \caption{Progressive learning for LA scar segmentation from LGE MRI. In Stage I, the model learns to identify the easy LA structure; in Stage II, the model inherited from Stage I, and learns to identify LA and scar at the same time with the use of two decoders as well as LA wall constraint for scar; in Stage III, the model inherited from Stage II, and keeps fine-tuning scar segmentation.}
    \label{fig:training}
\end{figure}

\subsection{Progressive learning framework}

We propose a three-stage progressive learning strategy for LA scar segmentation, in which the model gradually transitions from coarse anatomical localization to fine-grained pathological tissue detection. Figure \ref{fig:training} illustrates the overall pipeline.

\vspace{6pt}\noindent\textbf{Stage I: Pre-training on LA segmentation.}
In the first stage, the model is randomly initialized and trained exclusively on LA cavity segmentation task to establish anatomical priors. Given an augmented input LGE MRI volume $\mathcal{X'}$, the encoder of DL model $\mathcal{F}$ extracts LA features, which are then processed by the decoder to produce the LA mask $\mathcal{\hat{Y}}_{LA}^{(\text{I})}$. They are defined as:

\begin{equation}
\begin{cases}
\mathcal{\hat{Y}}_{\text{LA}}^{(\text{I})} &= \mathcal{F}(\mathcal{X'}; \theta^{(0)}) \\
\mathcal{L}^{(\text{I})} &= \mathcal{L}_{\text{DiceCE}}(\mathcal{\hat{Y}}_{\text{LA}}^{(\text{I})}, \mathcal{Y}_{\text{LA}}) \\
\theta^{(\text{I})} &\leftarrow \mathcal{O}(\theta^{(0)}, \nabla_{\theta} \mathcal{L}^{(\text{I})})
\end{cases}
\end{equation}

where $\mathcal{Y}_{\text{LA}}$ denotes ground truth LA mask. The network parameters ($\theta$) are updated by minimizing DiceCE loss ($\mathcal{L}^{(\text{I})}$) to maintain stable anatomical representation learning, via backpropagation using the AdamW optimizer ($\mathcal{O}$). This stage focuses on learning the overall cardiac anatomy without the complexity of scar detection, avoiding training instability from simultaneous multitask optimization from scratch.

\vspace{6pt}\noindent\textbf{Stage II: Dual-task learning.}
Building on the pre-trained LA segmentation model, the second stage introduces a parallel scar segmentation branch while maintaining LA prediction as a regularization task. The architecture adopts a dual-decoder design where the encoder ($\mathcal{E}$) and LA decoder ($\mathcal{D}_{\text{LA}}$) inherit pretrained weights from Stage I, and the newly introduced scar decoder ($\mathcal{D}_{\text{scar}}$) is randomly initialized. Both decoders share the same representations ($\mathcal{Z}$) from pretrained encoder. Formally:

\begin{equation}
\begin{cases}
\mathcal{Z} &= \mathcal{E}(\mathcal{X'}; \theta^{(\text{I})}) \\
\mathcal{\hat{Y}}_{\text{LA}}^{(\text{II})} &= \mathcal{D}_{\text{LA}}(\mathcal{Z};{\theta}^{(\text{I})}) \\
\mathcal{\hat{Y}}_{\text{scar}}^{(\text{II})} &= \mathcal{D}_{\text{scar}}(\mathcal{Z};{\theta^{(0)}}) \\
\end{cases}
\end{equation}


In this second stage, we adopt a dual-task learning strategy to optimize LA cavity and scar segmentation jointly. The overall loss is defined as:

\begin{equation}
    \mathcal{L}^{(\text{II})} = \beta_{\text{LA}}\mathcal{L}_{\text{LA}} + \beta_{\text{scar}}\mathcal{L}_{\text{scar}}
\end{equation}

where $\beta_{\text{LA}}=0.3$ and $\beta_{\text{scar}}=0.7$ are the task weighting coefficients, emphasizing scar segmentation as the primary objective in this stage. LA and scar segmentation loss ($\mathcal{L}_{\text{LA}}$, $\mathcal{L}_{\text{scar}}$) is the same as $\mathcal{L}^{(\text{I})}$, which is DiceCE loss. 

To incorporate anatomical priors, we further introduce a soft LA wall constraint. Specifically, a wall mask ($\mathcal{M}$) is constructed using Euclidean distance transforms from the inner and outer LA boundaries. Voxels within predefined inner and outer thickness thresholds are considered part of the atrial wall, which is defined as follows: 
\begin{equation}
    \mathcal{M}_{\text{wall}} = \textbf{1}(d_{\text{in}} \le\delta_{\text{in}} \wedge d_{\text{out}} \le\delta_{\text{out}})
\end{equation}

where $\textbf{1}(\cdot)$ is indicator function, $\delta_{\text{in}}$ defines the thickness extending inside the LA cavity, and $\delta_{\text{out}}$ defines the thickness extending outside LA wall. The wall thicknesses are set to $\delta_{\text{in}}=3.0$mm and $\delta_{\text{out}}=2.5$mm based on empirical analysis on the scar distribution in LASCARQS public dataset. A spatial weight map ($\mathcal{W}$) is then defined as:

\begin{equation}
    \mathcal{W} = 1+ \alpha(t) \mathcal{M}_{\text{wall}}
\end{equation}

where the weighting factor $\alpha(t)$  is gradually increased during training to avoid imposing strong anatomical constraints before the model has learned reliable scar appearance features. Since the wall prior is only used as a training regularization factor rather than a test-time constraint, ground-truth LA annotation is employed to provide a stable and noise-free anatomical guidance. Scar loss is computed under this spatial weighting scheme to enforce scar predictions within anatomically plausible wall regions. The weighted scar loss is computed as:

\begin{equation}
\mathcal{L}_{\text{scar}} = \ell_{\text{scar}}(\mathcal{W}\odot\hat{\mathcal{Y}}_{\text{scar}}^{(\text{II})},
\mathcal{W}\odot\mathcal{Y}_{\text{scar}}^{(\text{II})})
\end{equation}

where $\ell_{\text{scar}}$ stands for binary segmentation loss. Both the prediction and the ground-truth label are multiplied by the same spatial weighting matrix to incorporate anatomical prior knowledge without altering the fundamental form of the segmentation loss.

Clinically, post-ablation scar is known to occur exclusively within LA wall. However, manual annotations may include occasional labeling noise outside the wall region or inside LA cavity (see Figure \ref{fig:lascar}). Therefore, the proposed weighting scheme serves as a soft spatial constraint that suppresses anatomically implausible outlier scars and prevents the model from learning such annotation-based bias.

Importantly, when scar is correctly confined within LA wall, the weighting matrix reduces to a spatially constant factor, and the loss is equivalent to standard segmentation loss. Thus, the proposed formulation preserves the original optimization objective while improving anatomical robustness. This may slightly reduce quantitative evaluation, since some scar ground-truth annotations located outside LA wall are intentionally down-weighted. However, this design improves anatomical consistency and enhances model generalization by preventing overfitting to annotation noise.

To maintain anatomical priors learned in Stage I, we employ a differentiated parameter update strategy: the encoder and LA decoder are updated cautiously to preserve their representations, whereas the scar decoder is trained with a standard update schedule to learn scar segmentation.

\vspace{6pt}\noindent\textbf{Stage III: LA scar segmentation fine-tuning.} This stage focuses exclusively on scar learning, allowing the network to adapt to the specific scar segmentation task while retraining anatomical priors learned in previous stages. Compared to Stage II, the network is simplified, as only the scar decoder and a reduced set of trainable parameters are actively fine-tuned. Given the input LGE MRI volume $\mathcal{X'}$, the network $\mathcal{F}(\cdot; \theta^{(\text{II})})$ predicts the scar segmentation map $\hat{\mathcal{Y}}_{\text{scar}}^{(\text{III})}$:

\begin{equation}
\begin{cases}
\mathcal{\hat{Y}}_{\text{scar}}^{(\text{III})} &= \mathcal{F}(\mathcal{X'}; \theta^{(\text{II})}) \\
\mathcal{L}^{(\text{III})} &= \mathcal{L}_{\text{DiceCE}}(\mathcal{\hat{Y}}_{\text{scar}}^{(\text{III})}, \mathcal{Y}_{\text{scar}}) \\
\theta^{(\text{III})} &\leftarrow \mathcal{O}(\theta^{(\text{II})}, \nabla_{\theta} \mathcal{L}^{(\text{III})})
\end{cases}
\end{equation}

where $\mathcal{Y}_{\text{scar}}$ denotes ground-truth scar mask. The parameters $\theta^{(\text{II})}$ inherited from Stage II are updated to $\theta^{(\text{III})}$ via an optimizer ($\mathcal{O}$). During this stage, the early encoder parameters are frozen to preserve low level anatomical representations learned during pretraining and to mitigate catastrophic forgetting during scar-specific fine-tuning.

\section{Experiments}

\subsection{Dataset and preprocessing}

\begin{figure}[t]
\centering
\begin{subfigure}{0.24\linewidth}
    \centering
    \includegraphics[width=\linewidth]{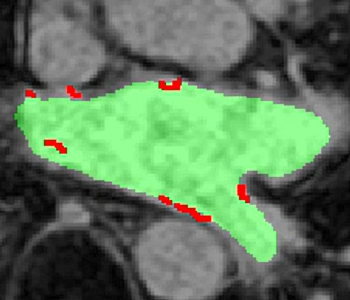}
\end{subfigure}
\begin{subfigure}{0.24\linewidth}
    \centering
    \includegraphics[width=\linewidth]{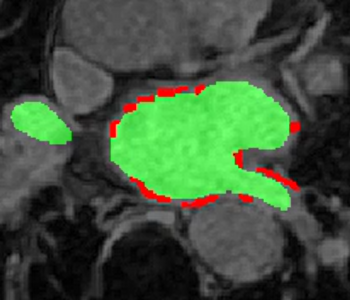}
\end{subfigure}
\begin{subfigure}{0.24\linewidth}
    \centering
    \includegraphics[width=\linewidth]{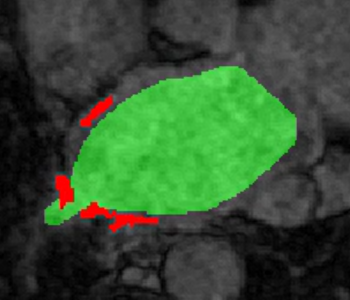}
\end{subfigure}
\begin{subfigure}{0.24\linewidth}
    \centering
    \includegraphics[width=\linewidth]{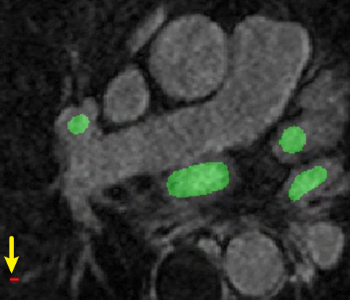}
\end{subfigure}
\caption{Examples of cropped axial images from LGE MRI volumes along with LA cavity (green) as well as scar (red) expert annotations from LASCARQS public dataset.}
\label{fig:lascar}
\end{figure}

We used the dataset from LASCARQS 2022 challenge \cite{zhuang2023left}, which included two subsets: subset 1 comprised $n = 60$ 3D LGE MRI with the corresponding LA cavity and scar expert annotations, while subset 2 comprised $n = 130$ 3D LGE MRI with only LA cavity annotations. Data were collected from 3 centers, including both pre-ablation and post-ablation images acquired in patients with AF, with two types of voxel size, ($0.625\times0.625\times2.5$) and ($1.0\times1.0\times1.0$) mm$^3$. In clinical practice, post-ablation scar is typically located within the LA wall surrounding the cavity. However, in this dataset, scar annotations were observed both inside and outside the LA cavity, which violates anatomical plausibility. Such inconsistencies may introduce label noise and adversely affect segmentation model training stability and generalization performance. Examples are shown in Figure \ref{fig:lascar}. We preprocessed data by resampling all images to the same pixel spacing ($0.625\times0.625\times2.5$) mm$^3$ and cropping to the same size ($384\times256\times40$) voxels. The training patch size was ($228\times192\times32$), Z-score normalization and data augmentation were applied during training.

\subsection{Experimental protocol}

Our training framework included 3 stages, in stage I, preprocessed dataset included 130 3D LGE MRI with LA annotations; in Stage II and III, the preprocessed dataset has 60 3D LGE MRI, with LA and scar annotations. It should be noted that the method was validated on a single small public dataset. External validation was lacking since we did not have access to validation/test sets. Therefore, model performance was evaluated using an internally split validation set derived from the publicly available training data. During each stage, 20\% of the data was used as validation, 5-fold cross-validation was performed. The evaluation metrics were Dice score (Dsc), Hausdorff Distance (HD) and Average Surface Distance (ASD). We set a maximum of 250 training epochs, and early stopping was employed based on the validation Dsc with a patience of 30 epochs and a minimum improvement threshold of 0.0001. The backbone was SwinUNETR model \cite{hatamizadeh2021swin}, which has a U-shaped network design and uses a Swin transformer \cite{liu2021swin} as the encoder and CNN-based decoder that is connected to the encoder via skip connections at different resolutions. The model had 62.6 million trainable parameters. The model was trained from scratch in Stage I. Learning rate was 1e-4. Batch size was 1. The optimizer was AdamW.  The training plan was implemented using mainly PyTorch and MONAI library and training was performed on dual Intel\textsuperscript{\textregistered} Xeon\textsuperscript{\textregistered} Gold 6226R 3.90 GHz (16 cores), 512 GB ddr4 RAM, NVIDIA RTX A6000 GPU, 48 GB of dedicated memory. To ensure reproducibility and comparability of results, the same fixed random seed was used for each training session.

\subsection{Results}

\begin{figure}[ht]
    \centering
    \includegraphics[width=0.95\linewidth]{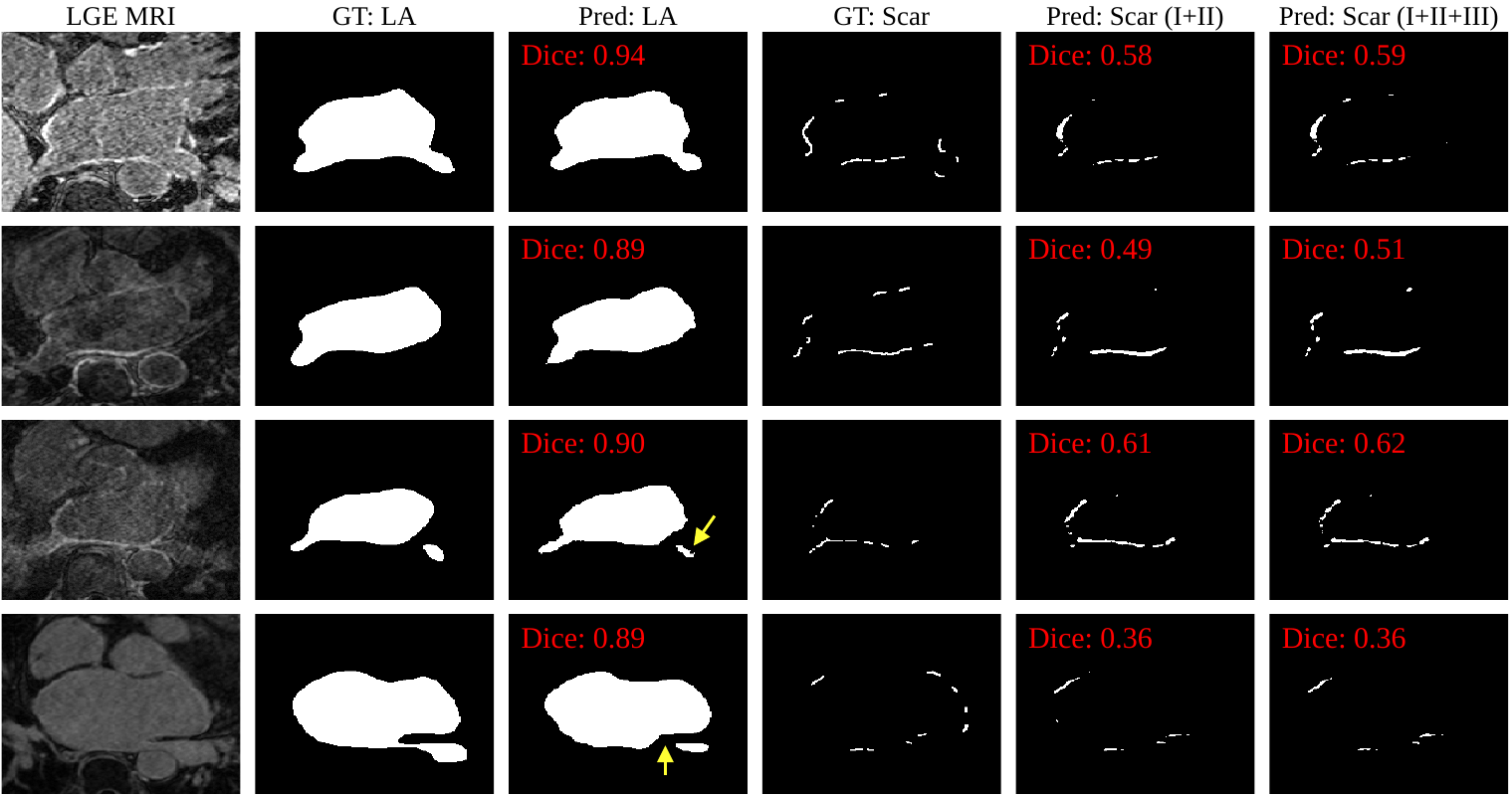}
    \caption{Examples of LA cavity and scar segmentation from LASCARQS public dataset. From left to right columns: input LGE MRI images, LA ground truth (GT), LA predicitons from Stage I+II. LA scar GT, scar predictions from Stage I+II and Stage I+II+III respectively. The Dice scores are indicated in red.}
    \label{fig:results}
\end{figure}

Figure \ref{fig:results} provides examples of LA and scar segmentation results in comparison with ground truth. The overall anatomical shape was well preserved, and predicted boundaries closely matched ground truth in most regions. However, the model still struggled in thin and complex connection areas, where slight discontinuities or boundary inaccuracies can be observed. 

\begin{table}[t]
\centering
\caption{Performance for LA and scar segmentation on validation set in terms of Dice score (Dsc) $\uparrow$, Hausdorff distance (HD) $\downarrow$ and Average Surface Distance (ASD) $\downarrow$.}
\label{tab:results}

\begin{tabularx}{\textwidth}{XXXX}
\toprule
\multicolumn{4}{c}{{LA cavity segmentation ($n = 12$)}} \\
\midrule
Stage & Dsc (\%) & HD (mm) & ASD (mm) \\
\midrule
I & $\textbf{0.94}\pm\textbf{0.02}$ & $\textbf{3.05}\pm\textbf{0.96}$ & $\textbf{0.78}\pm\textbf{0.32}$ \\
I+II & $0.90\pm0.01$ & $8.30\pm2.44$ & $2.28\pm1.13$ \\
I+II+III & -- & -- & -- \\
\midrule
\multicolumn{4}{c}{{LA scar segmentation ($n = 12$)}} \\
\midrule
Stage & Dsc (\%) & HD (mm) & ASD (mm) \\
\midrule
I & -- & -- & -- \\
I+II & $0.49\pm0.08$ & $13.02\pm9.15$ & $1.96\pm1.33$ \\
I+II+III & $\textbf{0.50}\pm\textbf{0.02}$ & $\textbf{11.84}\pm\textbf{3.11}$ & $\textbf{1.80}\pm\textbf{0.55}$ \\
\bottomrule
\end{tabularx}
\end{table}

Table \ref{tab:results} summarizes performances of the proposed framework for LA and scar region segmentation under different training stage combinations. For LA segmentation, the model had robust performance with Dsc of 0.94 in Stage I, besides, we observed a mild degradation in LA segmentation performance when emphasizing scar optimization, Dsc decreased to 0.90 in Stage II, HD and ASD presented increased variability, as reflected by their larger standard deviations compared with Stage I. This behavior may be attributed to gradient interference in the shared encoder of the multi-task architecture, where scar-driven optimization influenced the representation learning for LA segmentation. Furthermore, the asymmetric loss weighting strategy ($\beta_{\text{LA}}=0.3$ versus $\beta_{\text{scar}}=0.7$) biased the optimization process toward scar prediction.
In the full model (Stage I+II+III) setting, scar segmentation achieved Dsc of 0.50, increased by 2\% compared to the model of Stage I+II, and the HD, ASD were also reduced by 9\% and 8\%, respectively. 
This result demonstrated that while the dual-task learning in Stage II provides anatomical priors, extended joint optimization led to task dominance, biasing the model toward LA segmentation and consequently degrading performance in small lesion (i.e., scar) segmentation performance. Hence, performing the subsequent single-task fine-tuning appears useful. While slight but consistent improvement was brought by Stage I + II + III compared to Stage I + II, indicating effectiveness and rationality of the proposed design, Dsc for scar segmentation remained relatively low in both settings. This suggested that the model faces inherent challenges when predicting extremely small, fragmented, and irregular targets, which are characterized by weak contrast, high variability, and limited spatial continuity.
Compared with previously reported methods \cite{zhuang2023left}, in which the best Dsc of scar on test set was 0.595, our approach did not demonstrate a clear performance improvement. This is due to differences in validation data and experiment protocols, a direct quantitative comparison with previous challenge results is not feasible.

\subsection{Ablation study}

To evaluate the effectiveness of each training stage for LA scar segmentation in the proposed framework, we conducted an ablation study, as summarized in Table \ref{tab:ablation}. Baseline 1 (B1) including only Stage III training, i.e., direct supervised learning without any pretraining. This setting achieved a mean Dsc of 0.49, HD of 14.16 mm, and ASD of 2.12 mm. Baseline 2 (B2) incorporated Stage I pretraining followed directly by Stage III fine-tuning, without intermediate Stage II dual-task transition. In this case, the Dsc improved to 0.50, along with a decrease in HD and ASD to 11.67 mm and 1.72 mm, respectively, demonstrating consistent improvements over B1. This experiment confirmed that representation learning on LA structures alone was less performant to segment scar without gradually introducing scar-specific supervision, and that anatomical priors learned from LA segmentation effectively facilitated scar learning.
Baseline 3 (B3) trained only Stage II without Stage I pretraining to assess importance of supervised LA initialization. Baseline 4 (B4) combined Stage II (trained from scratch) with Stage III. Comparisons between B3 and B4 suggested that both Stage I and Stage II had an added value in the training pipeline.
Furthermore, due to the soft wall constraint introduced in Stage II, some annotated scar regions located outside the LA wall (those are anatomically implausible) were down-weighted during optimization, which inevitably led to slightly lower quantitative metrics. Nevertheless, this design improved anatomical consistency and contributed to the overall effectiveness of the complete training strategy (Stage I + II + III).

We also evaluated nnU-Net \cite{isensee2021nnu}, a widely adopted self-configuring framework that serves as a strong and standardized baseline for medical image segmentation, on the same dataset, yielding mean Dsc of 0.50, HD of 16.10 mm, and ASD of 1.51 mm, highlighting the challenging nature of scar segmentation. In terms of computational cost, nnU-Net required approximately 3 hours per fold (about half of our training time), while SwinUNETR consumed more GPU memory and time due to its transformer-based architecture. Because of its small size and highly fragmented distribution, LA scar inherently limit the achievable Dsc compared with larger anatomical structures such as the LA. From a clinical perspective, scar identification is not assessed through strict voxel-wise overlap, as inter-observer agreement relies more on regional interpretation. Therefore, the feasibility and clinical relevance of Dice as a primary evaluation metric for scar segmentation warrant careful consideration. 

\begin{table}[t]
\centering
\caption{Scar segmentation performances on validation set according to different baselines (B), in terms of Dsc, HD and ASD.}
\label{tab:ablation}
\begin{tabularx}{\textwidth}{XXXXXX}

\toprule
Baseline &{Stage} & Dsc (\%) $\uparrow$ & HD  (mm) $\downarrow$ & ASD (mm) $\downarrow$ \\
\midrule
nnUNET  & -- & $0.50\pm0.09$ & $16.10\pm24.33$ & $1.51\pm3.63$  \\
B1 & III & $0.49\pm0.02$ & $14.16\pm4.26$ & $2.12\pm0.76$  \\
B2 & I+III & $0.50\pm0.02$ & $11.67\pm2.48$ & $1.72\pm0.34$ \\
B3$^*$ & II &  $0.45\pm0.02$ & $14.49\pm9.86$ & $2.98\pm1.71$ \\
B4$^*$ & II+III &  $0.49\pm0.01$ & $11.25\pm0.80$ & $1.84\pm0.81$ \\
Full & I+II+III &  $0.50\pm0.02$ & $11.84\pm3.11$ & $1.80\pm0.55$ \\
\bottomrule
\end{tabularx}%
\par\smallskip
\raggedright\noindent{\footnotesize $^*$ Model parameters in Stage II were initialized randomly.}
\end{table}

\section{Conclusion}
\label{sec:conclusion}
In this work, we propose a progressive learning framework to quantify LA scar distribution from LGE MRI images.  Given the limited size and imperfect annotation \cite{lefebvre2022lassnet} quality of available datasets, we further incorporated clinically meaningful data augmentation and introduced a LA wall constraint to enforce anatomical plausibility during training. This progressive strategy is particularly suitable for tasks where direct optimization is difficult, allowing the model to first learn global anatomy before focusing on subtle pathological patterns. In future studies, we aim to further refine and validate our approach on local LGE MRI data from AF patients, integrating both AI-based analysis and clinical expertise.

\section*{Acknowledgement}

This research is funded by Institut Universitaire d’Ingénierie en Santé (IUIS).

\bibliography{reference.bib}
\bibliographystyle{unsrt}

\end{document}